\documentclass[10pt]{article}

\usepackage[letterpaper, textwidth=5.5in, textheight=9in, centering, headheight=12pt, headsep=18pt]{geometry}
\usepackage[T1]{fontenc}
\usepackage{mathpazo}
\usepackage{inconsolata}
\usepackage{microtype}
\usepackage{needspace}
\usepackage{placeins}
\usepackage{float}
\usepackage{amsmath}
\usepackage{booktabs}
\usepackage{graphicx}
\usepackage{xcolor}
\usepackage{tikz}
\usetikzlibrary{arrows.meta,positioning}

\usepackage{titlesec}
\titleformat{\section}{\large\bfseries}{\thesection}{1em}{}
\titleformat{\subsection}{\normalsize\bfseries}{\thesubsection}{1em}{}
\titlespacing*{\section}{0pt}{14pt}{6pt}
\titlespacing*{\subsection}{0pt}{10pt}{4pt}
\titleformat{\paragraph}[runin]{\normalsize\bfseries}{}{0pt}{}
\titlespacing*{\paragraph}{0pt}{6pt}{0.6em}

\usepackage{fancyhdr}
\usepackage[round,authoryear]{natbib}
\definecolor{linkblue}{RGB}{56,118,196}
\usepackage[colorlinks=true, citecolor=linkblue, linkcolor=linkblue, urlcolor=linkblue]{hyperref}

\usepackage{caption}
\newcommand{\nAll}{86{,}870}
\newcommand{\nBestFoneStep}{4{,}000}
\newcommand{\nBestFoneVal}{62.1}
\newcommand{\nBestStep}{5{,}000}

\newcommand{\nDiffLRfCwe}{+3.7}
\newcommand{\nDiffLRfCweHi}{+4.0}
\newcommand{\nDiffLRfCweLo}{+3.4}
\newcommand{\nDiffLRfFone}{+4.1}
\newcommand{\nDiffLRfFoneHi}{+5.2}
\newcommand{\nDiffLRfFoneLo}{+3.0}
\newcommand{\nDiffLRfMae}{-0.09}
\newcommand{\nDiffLRfMaeHi}{-0.08}
\newcommand{\nDiffLRfMaeLo}{-0.10}
\newcommand{\nDiffLRfSev}{+2.8}
\newcommand{\nDiffLRfSevHi}{+3.1}
\newcommand{\nDiffLRfSevLo}{+2.4}
\newcommand{\nDiffLRfVec}{+4.7}
\newcommand{\nDiffLRfVecHi}{+5.0}
\newcommand{\nDiffLRfVecLo}{+4.3}
\newcommand{\nDropped}{561}
\newcommand{\nDroppedPct}{1.4}

\newcommand{\nEitherSev}{75.8}

\newcommand{\nExPanOSPred}{9.8}
\newcommand{\nExPanOSTrue}{8.8}
\newcommand{\nExPurityPred}{9.8}
\newcommand{\nExPurityTrue}{7.2}

\newcommand{\nExRoboDKTrue}{3.3}

\newcommand{\nLRBfRecLow}{19.2}

\newcommand{\nLRdBias}{+0.08}
\newcommand{\nLRdCritPrec}{52.1}

\newcommand{\nLRdFone}{47.9}
\newcommand{\nLRdFoneCritical}{51.4}

\newcommand{\nLRdFoneLow}{2.2}

\newcommand{\nLRdMae}{0.98}

\newcommand{\nLRdMaeLow}{2.91}

\newcommand{\nLRdRecCritical}{50.7}

\newcommand{\nLRfBias}{+0.33}

\newcommand{\nLRfCompC}{82.9}
\newcommand{\nLRfCompI}{82.9}
\newcommand{\nLRfCompPR}{76.5}

\newcommand{\nLRfCritPrec}{45.4}
\newcommand{\nLRfCritPred}{7{,}755}
\newcommand{\nLRfCritTrue}{4{,}800}

\newcommand{\nLRfFone}{51.6}
\newcommand{\nLRfFoneCritical}{56.1}

\newcommand{\nLRfFoneLow}{10.9}

\newcommand{\nLRfMae}{0.94}

\newcommand{\nLRfNineEight}{17.7}

\newcommand{\nLRfRecCritical}{73.3}

\newcommand{\nLRfSev}{67.1}
\newcommand{\nLRfSevQA}{59.7}

\newcommand{\nLRfSoleAvail}{16}

\newcommand{\nLRfSubSev}{65.3}

\newcommand{\nLRfVecOne}{28.1}

\newcommand{\nLRfVecThreePlus}{13.6}
\newcommand{\nLRfVecTwo}{14.1}
\newcommand{\nLRfVecZero}{44.1}

\newcommand{\nLenBB}{194}
\newcommand{\nLenBC}{285}
\newcommand{\nLenBD}{444}

\newcommand{\nLoRAfBias}{+0.33}

\newcommand{\nLoRAfCompC}{84.8}
\newcommand{\nLoRAfCompI}{84.5}
\newcommand{\nLoRAfCompPR}{78.8}

\newcommand{\nLoRAfCritPrec}{48.7}
\newcommand{\nLoRAfCritPred}{7{,}568}
\newcommand{\nLoRAfCritTrue}{4{,}800}
\newcommand{\nLoRAfCwe}{70.9}

\newcommand{\nLoRAfMaeCritical}{0.53}
\newcommand{\nLoRAfMaeHigh}{0.76}
\newcommand{\nLoRAfMaeLow}{2.90}
\newcommand{\nLoRAfMaeMedium}{0.92}
\newcommand{\nLoRAfNineEight}{20.0}

\newcommand{\nLoRAfRareCwe}{7.0}

\newcommand{\nLoRAfRecLow}{11.4}

\newcommand{\nLoRAfSev}{69.9}
\newcommand{\nLoRAfSevQA}{61.7}

\newcommand{\nLoRAfSevQC}{74.4}
\newcommand{\nLoRAfSevQD}{73.9}
\newcommand{\nLoRAfSoleAvail}{17}
\newcommand{\nLoRAfSolePR}{38}

\newcommand{\nLoRAfSubSev}{69.6}

\newcommand{\nLoRAfVec}{48.8}
\newcommand{\nLoRAfVecOne}{25.5}

\newcommand{\nLoRAfVecThreePlus}{13.0}
\newcommand{\nLoRAfVecTwo}{12.7}
\newcommand{\nLoRAfVecZero}{48.8}
\newcommand{\nLoraParams}{12.6}
\newcommand{\nLoraPct}{0.329}
\newcommand{\nLoraTestHours}{4.9}

\newcommand{\nLoraTrainHours}{2.5}

\newcommand{\nNvdNineEight}{10.5}

\newcommand{\nRFdMae}{0.98}

\newcommand{\nRFdSev}{66.9}

\newcommand{\nRFfMae}{1.03}

\newcommand{\nRFfSev}{64.0}

\newcommand{\nRareCweN}{3{,}023}
\newcommand{\nSub}{2{,}000}
\newcommand{\nSubCritShare}{13.3}
\newcommand{\nSubMediumShare}{47.5}
\newcommand{\nTest}{36{,}325}
\newcommand{\nTestClipped}{1.4}
\newcommand{\nTestCwes}{512}
\newcommand{\nTfidfMinutes}{9}
\newcommand{\nTopCov}{91.7}
\newcommand{\nTrain}{40{,}436}
\newcommand{\nTrainCwes}{464}
\newcommand{\nTrainVal}{50{,}545}
\newcommand{\nUsableShareA}{85.9}
\newcommand{\nUsableShareB}{90.7}
\newcommand{\nUsableShareC}{94.4}
\newcommand{\nVThreeZero}{776}
\newcommand{\nVal}{10{,}109}
\newcommand{\nZSCompA}{22}
\newcommand{\nZSCompAC}{93}
\newcommand{\nZSCompAV}{84}
\newcommand{\nZSCompC}{47}

\newcommand{\nZSCritShare}{62.4}
\newcommand{\nZSInvalid}{11}
\newcommand{\nZSMediumCount}{9}
\newcommand{\nZSMediumShare}{0.4}
\newcommand{\nZSParse}{99.7}
\newcommand{\nZSScoreCons}{3.4}
\newcommand{\nZSSevCons}{37.0}

\newcommand{\nZSdMae}{1.58}

\newcommand{\nZSdSev}{20.8}

\newcommand{\nZSfMae}{2.08}

\newcommand{\nZSfSev}{43.5}

\begin{document}

\thispagestyle{fancy}
\vspace*{0.35in}
\begin{center}
{\Large\bfseries Predicting CVSS and CWE from CVE Descriptions\\with Fine-Tuned Small Language Models\par}
\vspace{0.3in}
{\bfseries Tara Dixit \hspace{0.45in} Jessie Dong \hspace{0.45in} Summer Li\par}
\vspace{6pt}
Stanford University\par
\vspace{2pt}
{\small\ttfamily \{taradixit, jessiedong, summerli\}@stanford.edu\par}
\end{center}
\vspace{0.15in}

\begin{center}{\large\bfseries Abstract}\end{center}
\vspace{-6pt}
\begin{quote}
Until April 2026, the National Vulnerability Database labeled nearly every published Common Vulnerabilities and Exposures (CVE) entry with a Common Vulnerability Scoring System (CVSS) v3.1 vector, a severity tier, and a Common Weakness Enumeration (CWE) weakness type.
Now, NIST enriches only the high priority CVEs, leaving many newly published vulnerabilities unlabelled with the classifiers that security teams use for patch triage.
In CVE enrichment, we predict the eight base components of CVSS v3.1 vector, base score, severity tier, and CWE from the vulnerability's textual description alone.
To prevent leakage between near-duplicate descriptions, all systems are trained only on CVEs published in 2022 and 2023.
Results are evaluated on all \nTest{} usable 2024 CVEs in the pinned NVD snapshot.
All scores are either directly predicted, or calculated from predicted vectors using the CVSS formula.
A fine-tuned 3.8B-parameter model Phi-3.5-mini, trained with LoRA in \nLoraTrainHours{} hours on a single Apple M5 Pro chip, achieves \nLoRAfSev\% accuracy on severity and \nLoRAfCwe\% on CWE.
It matches the official vector exactly for \nLoRAfVec\% of test CVEs, outperforming TF-IDF logistic regression by 2.8 to 4.7 points on all accuracy metrics based on zero-exclusion paired bootstrap intervals.
When we evaluate the base model with a zero-shot prompt, it achieves \nZSdSev\% accuracy on severity in a sample of \nSub{} CVEs.
In this same sample, its stated scores match its own vector precisely for only \nZSScoreCons\% of CVEs.
However, deriving scores from predicted vectors boosts the critical recall of TF-IDF logistic regression from \nLRdRecCritical\% to \nLRfRecCritical\% with no additional cost.
And since the errors of the fine-tuned model and TF-IDF affect different CVEs, at least one system matches the correct severity tier for \nEitherSev\% of test CVEs.

\end{quote}

\suppressfloats[t]
\section{Introduction}\label{sec:intro}

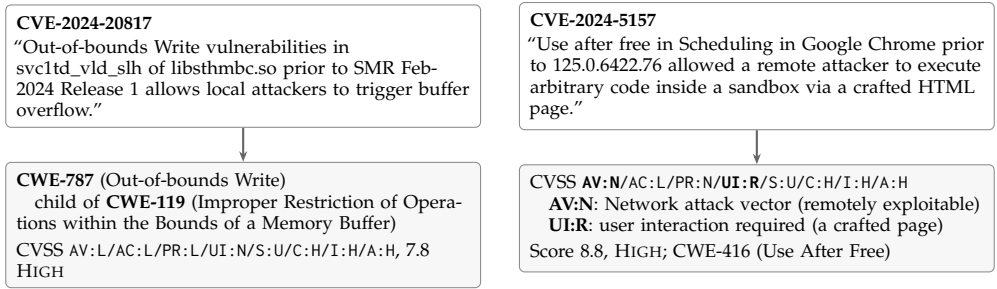
\begin{figure}[tb]
\centering
\begin{tikzpicture}[
  desc/.style={draw=black!45, rounded corners=2pt, align=left, font=\scriptsize, inner sep=4pt, text width=6.0cm},
  lab/.style={draw=black!45, fill=black!3, rounded corners=2pt, align=left, font=\scriptsize, inner sep=4pt, text width=6.0cm},
  arr/.style={-{Stealth[length=4pt]}, black!60, thick}]
\node[desc] (d1) {\textbf{CVE-2024-20817}\\[1pt]``Out-of-bounds Write vulnerabilities in svc1td\_vld\_slh of libsthmbc.so prior to SMR Feb-2024 Release 1 allows local attackers to trigger buffer overflow.''};
\node[lab, below=0.45cm of d1] (l1) {\textbf{CWE-787} (Out-of-bounds Write)\\ \quad child of \textbf{CWE-119} (Improper Restriction of Operations within the Bounds of a Memory Buffer)\\[2pt]
CVSS \texttt{AV:L/AC:L/PR:L/UI:N/S:U/C:H/I:H/A:H}, 7.8 \textsc{High}};
\draw[arr] (d1) -- (l1);
\node[desc, right=0.5cm of d1] (d2) {\textbf{CVE-2024-5157}\\[1pt]``Use after free in Scheduling in Google Chrome prior to 125.0.6422.76 allowed a remote attacker to execute arbitrary code inside a sandbox via a crafted HTML page.''};
\node[lab, below=0.45cm of d2] (l2) {CVSS \texttt{\textbf{AV:N}/AC:L/PR:N/\textbf{UI:R}/S:U/C:H/I:H/A:H}\\ \quad \textbf{AV:N}: Network attack vector (remotely exploitable)\\ \quad \textbf{UI:R}: user interaction required (a crafted page)\\[2pt]
Score 8.8, \textsc{High}; CWE-416 (Use After Free)};
\draw[arr] (d2) -- (l2);
\end{tikzpicture}
\caption{Two CVE descriptions from the 2024 test set and the labels NIST assigned to them. Left: an out-of-bounds write, CWE-787, which the CWE hierarchy places under the buffer-overflow class CWE-119. Right: a use-after-free in Chrome that is exploitable over the network once a user opens a crafted page.}
\label{fig:examples}
\end{figure}

Knowing the severity of a vulnerability and the kind of security property it violates allows one to triage the vulnerabilities to which their software is susceptible, and prioritize patches accordingly.
For example, suppose one is interested in whether their software is vulnerable to buffer overflows, as in the figure on the left.
The Common Weakness Enumeration (CWE) assigns each weakness a unique CWE-ID (e.g., CWE-120) and groups related CWEs hierarchically (e.g., buffer overflow weaknesses such as CWE-120 and CWE-787 are children of CWE-119), which one could leverage to answer this question.
Alternatively, suppose that one only wants to patch vulnerabilities that could lead to remote code execution (RCE) from untrusted websites, as in the figure on the right.
The Common Vulnerability Scoring System (CVSS) provides eight components to quantify how a vulnerability is exploited and its impact on Confidentiality, Integrity, and Availability (e.g., the Network Attack Vector component in the figure, meaning remotely exploitable), and combines these components into a single base score in [0, 10] and a severity tier to ease triage.\footnote{The eight components and score determine severity: $\le$3.9 \textsc{Low}, $\le$6.9 \textsc{Medium}, $\le$8.9 \textsc{High}, else \textsc{Critical}.}
These enrichments for Common Vulnerabilities and Exposures (CVE)\footnote{\url{https://www.cve.org}} identifiers are assigned by humans at the National Vulnerability Database (NVD) maintained by the National Institute of Standards and Technology (NIST).
However, due to not being able to keep up with the number of new vulnerability disclosures, a CVE backlog arose in early 2024.
As such, as of 15 April 2026, NIST will enrich only CVEs present in CISA's Known Exploited Vulnerabilities (KEV) catalog or affecting US Federal government or Executive Order (EO) 14028 critical software, and mark others as ``Lowest Priority -- not scheduled for immediate enrichment''~\citep{nist2026nvd}.
Consequently, it is important to study automated CVE enrichment.

In this paper, we pose CVE enrichment as the task of predicting the eight CVSS v3.1\footnote{Common Vulnerability Scoring System version 3.1.} components~\citep{first2019cvss31} shown in Figure~\ref{fig:examples}, as well as the CVSS score, severity tier, and CWE from only a text description of the vulnerability.
We consider CVEs from the last three years\footnote{Of the NVD records that were not rejected, \nUsableShareC\% of 2024 CVEs meet our inclusion criteria, against \nUsableShareB\% for 2023 and \nUsableShareA\% for 2022.} from the National Vulnerability Database, since CVEs recur, so training/test splits should account for time and train on older CVEs to avoid leakage.
Specifically, within a pinned NVD release, we train on the CVEs from 2022 and 2023, and test on the \nTest{} usable\footnote{Usable meaning containing required text description, enrichment ground truths, and having a CVSS v3.x (v3.0 or v3.1) base score.} 2024 CVEs.

\paragraph{CVE enrichment.} Given only the text description of a CVE vulnerability disclosure, recover its CVSS vector and thereby its CVSS score and severity tier, as well as its CWE.

We make the following contributions:
\begin{itemize}
  \item \textbf{Temporal benchmark for CVE enrichment.} We release a temporal benchmark for CVE enrichment free from leakage issues, based on a pinned snapshot of the NVD and validated with a reference CVSS calculator implementation. We acknowledge that our choice of ground truth, the NVD's vectors, are themselves the products of fallible human judgment: NVD analysts can and do disagree about correct vectors, reporting organizations can assign different scores to the same vulnerabilities, and NVD labels are revised after the fact. But without consensus there's no dataset, and by pinning a snapshot, at least our reported numbers won't change under us.
  \item \textbf{CVSS/CWE Benchmark.} We evaluate the effectiveness of (1) TF-IDF features with linear classification/regression, (2) a 3.8B-parameter Phi-3.5-mini LLM zero-shot, and (3) Phi-3.5-mini fine-tuned with LoRA at predicting (a) the eight CVSS components and deriving the score/severity from these, or (b) the CVSS score and CWE directly. We find that the fine-tuned Phi-3.5-mini achieves the best severity tier accuracy of the systems we evaluate, \nLoRAfSev\%,\footnote{Bonhomme and Dulaunoy (2025) report over 82\% severity accuracy with RoBERTa, but on a random split whose training data includes 2024 CVEs, so their numbers are not comparable to ours (Appendix~\ref{app:data}).} and CWE accuracy of \nLoRAfCwe\%. It matches the exact CVSS vector for \nLoRAfVec\% of test CVEs and outperforms the TF-IDF logistic regression baseline by \nDiffLRfSev{} to \nDiffLRfVec{} points across tasks ($^\ast$ statistically significantly w/ paired bootstrap resampling, intervals in Appendix~\ref{app:results} do not contain 0). We also find that the fine-tuned model and the TF-IDF baseline complement one another, erring on different CVEs: at least one gets severity correct for \nEitherSev\% of test CVEs. We provide more results and discuss insights, error cases, and implications in Sections~\ref{sec:findings} and~\ref{sec:discussion}.
  \item \textbf{Previous Work.} Earlier works have used bag-of-words representations with linear models to predict CVSS components and score~\citep{elbaz2020fighting}, BERT to predict CVSS components~\citep{shahid2021cvssbert}, RoBERTa to predict tier~\citep{bonhomme2025vlai}, and bag-of-words with random forests~\citep{aota2020automation} or RoBERTa~\citep{mosievskiy2026roberta} to predict CWE. Our benchmark expands this prior work by considering a strict temporal split, more recent LLMs, and joint prediction of CVSS and CWE. We provide more related work in Section~\ref{sec:related}.
  \item \textbf{Reproducible Code \& Artifacts.} We release our configs\footnote{Configs to fine-tune Phi-3.5-mini on CVSS and CWE and run all baselines on our datasets: \url{https://github.com/jessiedong01/cvss-cwe-prediction}.} and per-CVE predictions.\footnote{Predictions from all baselines for all test CVEs in JSON Lines format, where each line is a JSON object mapping field to the model's prediction.} Our code regenerates all numbers in this work given an NVD release; we use release \texttt{v2026.10.05-000016}.\footnote{Mirror hosted by the Fraunhofer Institute for Communication, Information Processing and Ergonomics (Fraunhofer FKIE)~\citep{fkie2026nvdfeeds}.}
\end{itemize}

\section{Our Evaluation}\label{sec:evaluation}

\paragraph{Dataset.}
CVE descriptions can be very repetitive: it is common to observe several near-duplicate descriptions of the same vulnerability (e.g., found in different packages), sometimes weeks or months apart.
Thus, a random split of the CVE data can produce a training set which is representative of the test data distribution.
In this work, we instead take a temporal approach: train only on 2022 and 2023 CVEs and test on 2024 CVEs, to determine how well each system can generalize to future vulnerabilities.
All CVE data is taken from a particular snapshot of a National Vulnerability Database (NVD) feed,\footnote{Release \texttt{v2026.10.05-000016} of the Fraunhofer FKIE mirror of the NVD JSON feeds~\citep{fkie2026nvdfeeds}.} so as to avoid discrepancies between statistics reported in this document and results of released artifacts.
We evaluate on all \nTest{} 2024 CVEs meeting the following inclusion criteria (see Appendix~\ref{app:stats} for statistics and additional information):
\begin{itemize}
  \item CVE has a CVSS v3.x base metric vector and severity label;
  \item CVE has a CWE that is specific (excluding the placeholders \texttt{NVD-CWE-noinfo} and \texttt{NVD-CWE-Other});
  \item CVE's NVD-provided description is in English and at least 20 characters long.
\end{itemize}

\paragraph{Approach and metrics.}
As discussed in the introduction, CVSS base score and severity are simple calculations~\citep{first2019cvss31}, functions of the eight different components.
Thus, given a vulnerability description, we want to train systems that either (1) predict a score, or (2) predict the eight components and then calculate the score from that vector~\citep{elbaz2020fighting,shahid2021cvssbert}, and analyze performance in terms of severity classification (accuracy and macro-averaged F$_1$-score), CWE classification (accuracy), and CVSS score and vector.
For CVSS score, we report mean absolute error (MAE), and for CVSS vector, we report the percent of CVEs for which all eight components are correctly predicted (``Exact CVSS'' in Table~\ref{tab:summary}).
We implemented a CVSS v3.1 calculator and verified it agrees with the CVSS base score listed by the NVD for all \nAll{} CVEs in our dataset.
Thus, any score errors reported are a result of mistakes in component prediction.

\paragraph{Systems evaluated.}
For CWE classification, CVSS score, vector, and severity classification we evaluate three systems: a TF-IDF + logistic regression classifier, an instruction-tuned 3.8B-parameter LLM prompted zero-shot for classification, and that same LLM after a short (\nLoraTrainHours{} hour) LoRA fine-tuning to predict CWE and CVSS.
TF-IDF-based classifiers are strong baselines on this task: \citet{talibzade2025llms} find TF-IDF features perform best for CWE and severity prediction, and find them challenging to outperform on NLP tasks using vulnerability description text.
Further details about data preparation and training can be found in Appendix~\ref{app:data}.

We release our code\footnote{\url{https://github.com/jessiedong01/cvss-cwe-prediction}} as well as our system's prediction for each CVE.\footnote{In the \texttt{results/} directory of the repository, one JSON Lines file per system.}
Statistics reported here have been updated to be calculated from a particular snapshot of an NVD feed~\citep{fkie2026nvdfeeds} release (\texttt{v2026.10.05-000016}); please run our provided scripts for calculation on this data release to exactly reproduce the numbers below and in Appendix~\ref{app:results}.

\section{Findings}\label{sec:findings}

\begin{table}[t]
\centering
\caption{Results on enriching test (2024) CVEs. Systems marked $^{\diamond}$ predict CVSS scores by predicting vectors and computing the score using the CVSS formula. For each scoring method, bold highlights the best performing system and underline indicates second best. See Table~\ref{tab:main} for all results.}
\label{tab:summary}
\small
\setlength{\tabcolsep}{4pt}
\begin{tabular}{lccccc}
\toprule
System & Severity acc. & Severity macro-F1 & CWE acc. & Score MAE $\downarrow$ & Exact CVSS \\
\midrule
TF-IDF + LR & \underline{67.1} & 47.9 & \underline{67.2} & \underline{0.98} & \underline{44.1} \\
TF-IDF + CB-LR & 65.7 & \underline{51.9} & \underline{67.2} & \underline{0.98} & 40.5 \\
TF-IDF + RF & 66.9 & 47.1 & 65.2 & \underline{0.98} & 41.8 \\
Fine-tuned Phi-3.5 & \textbf{69.9} & \textbf{55.7} & \textbf{70.9} & \textbf{0.85} & \textbf{48.8} \\
\midrule
TF-IDF + LR$^{\diamond}$ & \underline{67.1} & 51.6 & \underline{67.2} & \underline{0.94} & \underline{44.1} \\
TF-IDF + CB-LR$^{\diamond}$ & \underline{67.1} & \underline{53.9} & \underline{67.2} & 0.98 & 40.5 \\
TF-IDF + RF$^{\diamond}$ & 64.0 & 47.8 & 65.2 & 1.03 & 41.8 \\
Fine-tuned Phi-3.5$^{\diamond}$ & \textbf{69.9} & \textbf{55.7} & \textbf{70.9} & \textbf{0.85} & \textbf{48.8} \\
\bottomrule
\end{tabular}
\end{table}

\paragraph{Fine-tuned Phi-3.5 performs best on all tasks.}
Table~\ref{tab:summary} summarizes test set performance (all 2024 CVEs in our snapshot of the dataset that have required information present: $n = \nTest$).
We find our fine-tuned Phi-3.5 model achieves the best performance, outperforming the TF-IDF + LR system by \nDiffLRfSev{} to \nDiffLRfVec{} points on all tasks.

\paragraph{Exact CVSS.}
Here, we consider the full eight-component vector, which is predicted for every test CVE (see Appendix~\ref{app:exact}).
In general, our models often predict similar outputs and severity, but Table~\ref{tab:agreement} shows they are frequently in disagreement.
However, for \nEitherSev\% of CVEs, at least one model is correctly predicting severity.
We next examine where and how our models err in CVSS prediction.

\begin{table}[t]
\centering
\caption{Agreement between fine-tuned Phi-3.5 and TF-IDF logistic regression (formula scoring) on the 2024 test set (\%). The last column is the share of CVEs on which at least one of the two is correct.}
\label{tab:agreement}
\small
\begin{tabular}{lccccc}
\toprule
 & Both correct & Fine-tuned only & TF-IDF only & Neither & At least one \\
\midrule
Severity & 61.3 & 8.6 & 5.9 & 24.2 & 75.8 \\
CWE & 64.4 & 6.5 & 2.9 & 26.2 & 73.8 \\
Exact CVSS vector & 39.7 & 9.1 & 4.5 & 46.7 & 53.3 \\
\bottomrule
\end{tabular}
\end{table}

First, the fine-tuned model is biased towards assigning higher scores: in CVE-2024-0003, the vulnerability is rated Critical despite its CVSS vector needing higher privileges.
Similarly, RoboDK in CVE-2024-0257 is scored lower by NIST at \nExRoboDKTrue.
In both examples the vulnerability descriptions are terse, and do not clearly explain the conditions which reduced their final scores.
Our systems defaulted to the priors learned from similar weaknesses.

Overall, the fine-tuned model outperforms all TF-IDF systems, including ridge regression.
However, there is almost no difference in performance for the fine-tuned model whether the scores are directly predicted or calculated from predicted components using the CVSS 3.1 formula.
We confirm in Table~\ref{tab:consistency} that the scores and severities produced by the CVSS formula from the fine-tuned model's predicted components are in fact almost always identical to the directly generated values.
Thus, it is apparent from its training data, which conveniently lists the CVSS components before the score, that the fine-tuned model is already acting as though calculating scores using the formula and its generated vectors.
For the logistic regression model, using formula scoring actually improves performance: score MAE decreases from \nLRdMae{} to \nLRfMae{}, and macro-F1 increases from \nLRdFone\% to \nLRfFone\% (accuracy does not change at \nLRfSev\%).
This is partly due to the increased Critical recall (\nLRdRecCritical\% $\rightarrow$ \nLRfRecCritical\%) as seen in Table~\ref{tab:tiers}, resulting from predicting Critical for \nLRfCritPred{} CVEs in comparison to \nLRfCritTrue{} in the NVD.
As such, Critical F1 increases from \nLRdFoneCritical\% to \nLRfFoneCritical\%, along with a substantial improvement in Low F1 (\nLRdFoneLow\% $\rightarrow$ \nLRfFoneLow\%).
This comes at the cost of reduced Critical precision (\nLRdCritPrec\% $\rightarrow$ \nLRfCritPrec\%).
On the other hand, random forest's comparatively worse per component performance negatively influences its formula predicted scores (\nRFdSev\% $\rightarrow$ \nRFfSev\% in accuracy and \nRFdMae{} $\rightarrow$ \nRFfMae{} MAE).

\begin{table}[t]
\centering
\caption{Recall and F1 (\%) by true severity tier on the 2024 test set.}
\label{tab:tiers}
\small
\setlength{\tabcolsep}{3.5pt}
\begin{tabular}{lcccccccc}
\toprule
 & \multicolumn{2}{c}{Critical} & \multicolumn{2}{c}{High} & \multicolumn{2}{c}{Medium} & \multicolumn{2}{c}{Low} \\
\cmidrule(lr){2-3}\cmidrule(lr){4-5}\cmidrule(lr){6-7}\cmidrule(lr){8-9}
System & Rec. & F1 & Rec. & F1 & Rec. & F1 & Rec. & F1 \\
\midrule
TF-IDF + LR, direct & 50.7 & 51.4 & 62.2 & 61.1 & 77.9 & 77.1 & 1.1 & 2.2 \\
TF-IDF + LR, formula & 73.3 & 56.1 & 62.6 & 62.6 & 71.3 & 76.8 & 6.2 & 10.9 \\
TF-IDF + LR bal., formula & 57.1 & 52.8 & 61.9 & 62.2 & 75.6 & 77.2 & 19.2 & 23.2 \\
Phi-3.5 + LoRA, formula & 76.9 & 59.7 & 63.6 & 64.6 & 75.1 & 79.0 & 11.4 & 19.6 \\
\bottomrule
\end{tabular}
\end{table}

\begin{figure}[t]
\centering
\includegraphics[width=\linewidth]{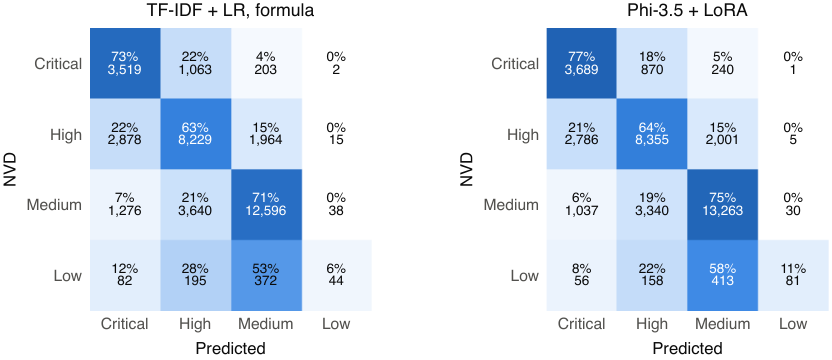}
\caption{Severity confusion matrices on the 2024 test set, normalized by true tier, for TF-IDF logistic regression with formula scoring (left) and fine-tuned Phi-3.5 (right).}
\label{fig:confusion}
\end{figure}

The Low severity tier is the most difficult for all systems, likely because it represents less than 2\% of our training data.
As shown in Figure~\ref{fig:confusion}, most other errors involve predicting adjacent severity levels.
It is encouraging that using class-balanced training weights greatly improves Low tier recall to \nLRBfRecLow\% in the TF-IDF model.
The fine-tuned model performed reasonably well in this under-represented tier (recall of \nLoRAfRecLow\%) even without data rebalancing.

\begin{figure}[t]
\centering
\includegraphics[width=0.62\linewidth]{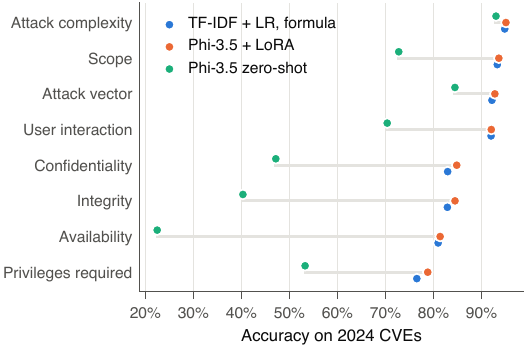}
\caption{Accuracy per CVSS component on 2024 CVEs. TF-IDF and the fine-tuned model are scored on the full test set, the zero-shot model on its \nSub{}-CVE sample.}
\label{fig:components}
\end{figure}

Figure~\ref{fig:components} shows per-CVSS-component accuracy on the 2024 test data, excluding the base score and severity which are dependent on the other components.
All models perform poorly on privileges required, with TF-IDF and the fine-tuned model at \nLRfCompPR\% and \nLoRAfCompPR\% respectively.
Attack complexity, scope, attack vector, and user interaction are likely easier to classify because they are almost always explicitly specified in the vulnerability description, and indeed, all systems perform well with above 90\% accuracy.
The next most difficult components after privileges required are the confidentiality, integrity, and availability impact scores, where \citet{marchiori2025canllms} found LLMs underperform embedding classifiers.
By contrast, we find that our fine-tuned model sees its largest improvements over TF-IDF on the less keyword dependent privileges required and CIA components, for instance, accuracy increases on the confidentiality and integrity components are \nLRfCompC\% $\rightarrow$ \nLoRAfCompC\% and \nLRfCompI\% $\rightarrow$ \nLoRAfCompI\% respectively.

\begin{figure}[t]
\centering
\includegraphics[width=0.62\linewidth]{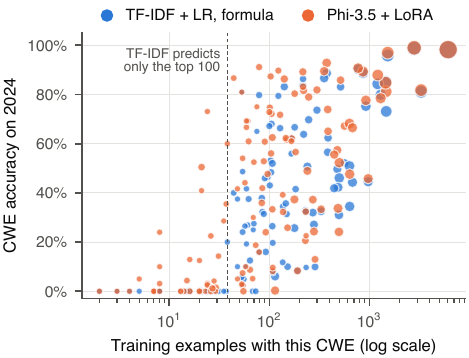}
\caption{CWE accuracy on the 2024 test set against the number of training examples with each CWE, for every CWE with at least 20 test CVEs. Point size grows with test count.}
\label{fig:cwe}
\end{figure}

Figure~\ref{fig:cwe} shows CWE accuracy on test set CVEs against how often each NVD-assigned CWE weakness appears in training.
Our TF-IDF model was limited to predicting on the 100 most common CWEs present in the training set, whereas the language model may predict an arbitrary textual CWE.
Accordingly, our TF-IDF model scores 0\% accuracy on the \nRareCweN{} CVEs with less commonly represented CWEs, while the fine-tuned model predicts \nLoRAfRareCwe\% correctly.
As expected, weaknesses that are common, and likely named explicitly in the vulnerability description, like cross-site scripting or SQL injection vulnerabilities, are near perfectly predicted by both of our trained models.
However, for less common weaknesses, or when the description is more general in nature, it appears our models learned associations between certain weaknesses and the severity metrics.

\begin{figure}[t]
\centering
\includegraphics[width=\linewidth]{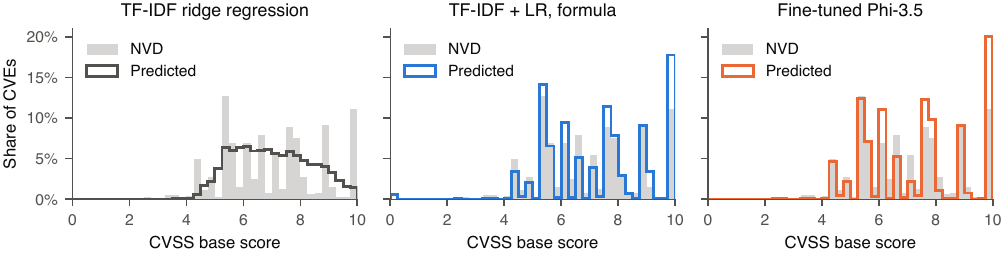}
\caption{Distribution of predicted and NVD base scores on the 2024 test set in bins of 0.25, for ridge regression (left), TF-IDF logistic regression with formula scoring (middle), and the fine-tuned model (right).}
\label{fig:scores}
\end{figure}

In Figure~\ref{fig:scores}, both TF-IDF and fine-tuned models over-predict the most common critical score more often than it is actually seen, likely also due to learned associations between weaknesses and common CVSS scores assigned to them.
By contrast, ridge regression predicts many scores unseen in the NVD data and does not predict the most critical or benign CVSS scores as frequently as they actually occur.
On the 2024 test set, logistic regression with formula scoring predicts a score of 9.8 (the most common critical score) on \nLRfNineEight\% of CVEs, in comparison to \nNvdNineEight\% in the NVD.
This bias results in a mean signed error of \nLRfBias, compared to only \nLRdBias{} for the ridge regression model which produces a more continuous score distribution.

To understand more deeply where and why models disagree with NVD-assigned vectors, we qualitatively analyzed errors.
Results are shown in Tables~\ref{tab:qualalla} through~\ref{tab:qualsomeb}: for each CVSS component, we chose four example CVEs from our 2024 test data where all (Tables~\ref{tab:qualalla} and~\ref{tab:qualallb}) or at least one (Tables~\ref{tab:qualsomea} and~\ref{tab:qualsomeb}) of our systems erred.
We bolded portions of the NVD description which may have served as clues for the true value.
In many cases only a single component differs between our predictions and the true value, and therefore exact match accuracy under-reports relative understanding of CVSS vectors.

We find the same general tendencies with the fine-tuned model as \citet{marchiori2025canllms} did with larger LLMs, namely, that some CVSS v3.1 components can be learned better than others.
Table~\ref{tab:valuefreq} compares the frequency of the possible values per CVSS v3.1 component that NIST assigned with those predicted by our model.
For example, the model overpredicts CVSS scores of 9.8 (\nLoRAfNineEight\% instead of \nNvdNineEight\%).
And correspondingly, it labels \nLoRAfCritPred{} of the CVEs from the test set as Critical, in comparison to \nLoRAfCritTrue{} CVEs labeled Critical in the NVD.
Thus, \nLoRAfCritPrec\% of those CVEs the model considers Critical are indeed Critical according to the NVD.
Overall, the model behaves similarly to the formula-scored logistic regression, meaning it is positively biased (mean signed error of \nLoRAfBias).

When analyzing the predicted vectors, we find that in \nLoRAfVecZero\% of the test CVEs, all vector components are predicted correctly (TF-IDF: \nLRfVecZero\%).
In \nLoRAfVecOne\% only one component is incorrect (TF-IDF: \nLRfVecOne\%), in \nLoRAfVecTwo\% two (TF-IDF: \nLRfVecTwo\%) and in \nLoRAfVecThreePlus\% at least three (TF-IDF: \nLRfVecThreePlus\%).
Thus, in many cases where our method misses, only one single detail is incorrectly predicted.
Often times, that detail may be hard to guess as it may be left out in the CVE description, like how exactly a DoS manifests itself (Conf.\ high or just avail.).
Also, the conditions for privilege requirements may be subtle to understand.
Overall, we can conclude that even the small fine-tuned language model significantly outperforms the classical baselines at learning how NIST enriches CVE descriptions with CVSS vectors, although the performance increase is not as big as the models sizes suggest it to be.

\begin{table}[t]
\centering
\caption{Comparison on the same random sample of 2{,}000 test CVEs, on which every system is scored.}
\label{tab:subset}
\small
\setlength{\tabcolsep}{4pt}
\begin{tabular}{lcccccc}
\toprule
 & \multicolumn{2}{c}{Severity} & CWE & Score & \multicolumn{2}{c}{Vector} \\
\cmidrule(lr){2-3}\cmidrule(lr){6-7}
System & Acc. & Macro-F1 & Acc. & MAE $\downarrow$ & Partial & Exact \\
\midrule
TF-IDF + LR, formula & 65.3\,{\scriptsize$\pm$2.1} & 47.6\,{\scriptsize$\pm$1.8} & 66.2\,{\scriptsize$\pm$2.2} & 0.99\,{\scriptsize$\pm$0.06} & 86.5\,{\scriptsize$\pm$0.7} & 42.2\,{\scriptsize$\pm$2.2} \\
Phi-3.5 zero-shot, direct & 20.8\,{\scriptsize$\pm$1.7} & 15.4\,{\scriptsize$\pm$2.4} & 25.7\,{\scriptsize$\pm$2.0} & 1.58\,{\scriptsize$\pm$0.05} & 60.5\,{\scriptsize$\pm$0.7} & 2.4\,{\scriptsize$\pm$0.6} \\
Phi-3.5 zero-shot, formula & 43.5\,{\scriptsize$\pm$2.3} & 26.8\,{\scriptsize$\pm$1.7} & 25.7\,{\scriptsize$\pm$2.0} & 2.08\,{\scriptsize$\pm$0.08} & 60.5\,{\scriptsize$\pm$0.7} & 2.4\,{\scriptsize$\pm$0.6} \\
Phi-3.5 + LoRA, formula & \textbf{69.6}\,{\scriptsize$\pm$2.0} & \textbf{52.8}\,{\scriptsize$\pm$3.3} & \textbf{68.8}\,{\scriptsize$\pm$2.1} & \textbf{0.87}\,{\scriptsize$\pm$0.05} & \textbf{87.5}\,{\scriptsize$\pm$0.7} & \textbf{48.0}\,{\scriptsize$\pm$2.1} \\
\bottomrule
\end{tabular}
\end{table}

Table~\ref{tab:subset} shows the predictions made by our zero-shot setup.
First of all, in \nZSParse\% of the cases the answer could be parsed as JSON object (from a random sample of \nSub{} CVEs, \nZSInvalid{} contained a valid JSON format but one of the CVSS components was invalid, e.g., a missing Scope value or Availability=CRITICAL).
We refer to Table~\ref{tab:zssev} for more details.
The table compares the distribution of CVSS severity written by the LLM to those that NIST derived for the same CVEs.
Clearly, the distribution is skewed towards Critical: \nZSCritShare\% of the CVEs are predicted Critical (NVD: \nSubCritShare\%) and accuracy for severity prediction is only \nZSdSev\%.
Medium is underpredicted (\nZSMediumCount{} CVEs, or \nZSMediumShare\%, instead of \nSubMediumShare\%).

Furthermore, we find that the written CVSS score is the result of the written CVSS vector in only \nZSScoreCons\% of the CVEs (see Table~\ref{tab:consistency}).
This may be due to the difficulty of performing arithmetic in context.
In fact, even for humans, CVSS calculation is considered difficult and tools exist to perform the computation.
We find similar problems when looking at the severity column of Table~\ref{tab:consistency}: here, we compare the severity label written by the LLM to the severity implied by the written score.
Only for \nZSSevCons\% they match.
Using the severity implied by the written vector instead increases the zero-shot severity accuracy on a random sample of \nSub{} test CVEs from \nZSdSev\% (using the written severity label) to \nZSfSev\%.
For comparison, on the same CVEs, fine-tuning achieves \nLoRAfSubSev\% and the TF-IDF method \nLRfSubSev\% severity accuracy.
The MAE of the scores worsens for zero-shot from \nZSdMae{} (using the written score) to \nZSfMae{} when instead calculating the score from the predicted CVSS vector.
Thus, the written scores are on average closer to the ground truth than those implied by the predicted CVSS vectors.

\begin{table}[htbp]
\centering
\caption{Self-consistency of language-model outputs (\%), where \emph{Score = formula} is the share of outputs whose written score equals the CVSS v3.1 score of the written vector and \emph{Severity = tier of score} is the share whose written severity matches the tier of the written score.}
\label{tab:consistency}
\small
\begin{tabular}{lccc}
\toprule
Model & Valid JSON & Score = formula & Severity = tier of score \\
\midrule
Phi-3.5 zero-shot & 99.7 & 3.4 & 37.0 \\
Phi-3.5 + LoRA & 100.0 & 99.8 & 100.0 \\
\bottomrule
\end{tabular}
\end{table}

\section{Discussion}\label{sec:discussion}

In the previous sections, we have already observed a couple of strengths and weaknesses in the predictions our methods make.
We further inspect common mistakes and compare our results.

\paragraph{Common Mistakes in CVSS Prediction.}
As can be seen in Figure~\ref{fig:components} and Tables~\ref{tab:qualsomea} and~\ref{tab:qualsomeb}, certain CVSS vector components are much easier to predict than others, no matter the method used for prediction.
Attack complexity and attack vector can be predicted with an accuracy of \nZSCompAV\% to \nZSCompAC\% zero-shot, whereas impact to availability, integrity and confidentiality are much more difficult to predict correctly (\nZSCompA\% to \nZSCompC\% zero-shot).

Furthermore, it becomes apparent from the tables that models tend to predict a CVSS vector for the weakness type mentioned in the CVE description if little further information is available or if the limiting factors are not mentioned or difficult to extract.

\paragraph{Training and Run Times.}
We find that severity classification accuracy (Table~\ref{tab:ckpt}) increases per checkpoint.
We validated on 1,000 CVEs from the validation split for each checkpoint (every 1,000 steps, up to 5,000 training steps).
While accuracy increases up to step \nBestStep, Macro-F$_1$ scores peak earlier (optimum at \nBestFoneStep{} steps with a score of \nBestFoneVal\%).
Depending on whether mispredictions of the rarer classes Critical and Low are considered more important to reduce than increasing overall accuracy, the model at step \nBestFoneStep{} might be preferred for deployment.

Fine-tuning takes \nLoraTrainHours{} h, scoring the test set (one year's worth of CVEs) another \nLoraTestHours{} h.
Batches contained at most 32 CVE descriptions.
For comparison, all TF-IDF experiments, including tuning, took only \nTfidfMinutes{} minutes.

Next, we investigate which CVSS vector components models struggle to predict correctly.
Table~\ref{tab:sole} presents the distribution of CVEs in which a given CVSS vector component is predicted incorrectly, for CVEs with one incorrect vector component.
We present this both for our fine-tuned model and for TF-IDF as a representative classical method.
Privileges Required makes up \nLoRAfSolePR\% of such examples for the fine-tuned model and TF-IDF respectively, with Availability Impact making up \nLoRAfSoleAvail\% and \nLRfSoleAvail\% respectively.

Finally, we investigate what causes the difference between CVEs with correct severity versus exact CVSS score match.
To our surprise, most of the gap can be attributed to CVEs with only one incorrect component (see Table~\ref{tab:gap}).
Thus, often a single ambiguous word can determine whether a given vulnerability is Critical or High, for example.
Two examples are presented in Table~\ref{tab:examples}.
In the case of CVE-2024-0003, it appears NIST inferred from the phrasing \emph{administrative service} that this is not easily exploitable by outsiders and gave it a score of \nExPurityTrue{} requiring high privileges.
However, the phrasing \emph{malicious user} led the model to predict an attacker does not need privileges, resulting in a Critical score of \nExPurityPred.
All other vector components were predicted correctly.
In the case of CVE-2024-0008, NVD gave a High score of \nExPanOSTrue{} as User Interaction was deemed required to exploit the vulnerability.
However, no mention is made of any user in the text, leading the model to make an incorrect prediction and give a score of \nExPanOSPred.
Here again, all other components were predicted correctly.
In both cases, one could argue that the model's interpretation is reasonable, given the missing details.

Next we examine MAE of the predicted score by true severity (Table~\ref{tab:tiermae}).
On the 2024 test set, our fine-tuned model's score MAE is \nLoRAfMaeCritical{} for Critical CVEs, \nLoRAfMaeHigh{} on High, \nLoRAfMaeMedium{} on Medium, and \nLoRAfMaeLow{} on Low CVEs.
Low severity CVEs are rare, making accurate prediction challenging for all methods.
Interestingly, Ridge without vector commitment achieves a score MAE of \nLRdMaeLow{} on Low CVEs.
When our model errs on a Low CVE it is usually by more than a severity tier.
Two examples are given in Table~\ref{tab:examples}.
In both cases, the descriptions contain words associated with particularly insidious weakness types.
The first CVE describes a heap overflow, a weakness type our model has likely seen associated with high severity during training, and makes a prediction accordingly.
However, the last sentence mitigates the severity, and NVD scored this accordingly.
The second CVE also vaguely alludes to a common weakness our model likely associates with high severity, where session-handling code can easily be construed as an authentication bypass vulnerability.
More information would be needed to clarify that this is likely less serious.
Thus, when the description alludes to a weakness that our model associates with serious vulnerabilities, but otherwise offers no relevant detail, the model appears to predict according to the weakness, rather than the sentence.
Low severity performs better for TF-IDF with class-balanced weights.
The analogue in a language model would be oversampling Low during training, which we did not explore.

We next analyze model performance as a function of the length of the CVE Description, split into quartiles 1\ldots4 from shortest to longest (with bounds \nLenBB, \nLenBC, and \nLenBD{} on the 2024 test set).
As shown in Table~\ref{tab:length}, model performance is lower when there is less text to work with (severity accuracy on the shortest CVEs being \nLoRAfSevQA\% and \nLRfSevQA\% for our fine-tuned model and TF-IDF respectively), with improvement and leveling off as description length increases (\nLoRAfSevQC\% on the third quartile of lengths and \nLoRAfSevQD\% on the fourth for our model, with a similar trend for exact vector).
Exact vector match results show the same trends as severity.
Many of the shortest descriptions are simply of the form PRODUCTNAME + weakness type, e.g.\ CVE-2024-0057 is titled \emph{.NET, .NET Framework, and Visual Studio Security Feature Bypass Vulnerability}, and the predicted score would be entirely dependent on the learned associations of Microsoft and SFBO vulnerabilities.
While the extra capacity of the larger fine-tuned model may be helpful overall, it has a similar advantage over TF-IDF in all quartiles, and thus does not appear to overcome poor performance on very short CVE descriptions.

\section{Related Work}\label{sec:related}

\paragraph{Automatic vulnerability triage.}
\citet{elbaz2020fighting} considered the problem of triaging new vulnerabilities as they are disclosed, and asked which triage decisions an analyst could automate using only the information available at that time.
For predicting the eight base CVSS metrics from vulnerability descriptions represented as bags of words, they found a linear classifier to perform best.
They report both directly predicting CVSS scores, and predicting CVSS vectors from which to compute CVSS scores.
CVSS-BERT~\citep{shahid2021cvssbert} takes a similar approach, training a BERT classifier for each metric, and use the specification's formula to compute CVSS scores from predicted vectors.
\citet{marchiori2025canllms} find that a classifier over strong embeddings outperforms general-purpose LLMs, at least on the three impact-related CVSS metrics.
VLAI~\citep{bonhomme2025vlai} fine-tunes RoBERTa on more than 600,000 vulnerabilities to predict the severity tier.
VLAI was trained on more than 600,000 vulnerabilities including CVEs from 2024 and later years, meaning we cannot fairly compare to this system on our test set.

For CWE prediction, \citet{aota2020automation} predict using random forest classification over bag-of-words features.
More recently, \citet{mosievskiy2026roberta} show that a small (125M parameter) RoBERTa language model fine-tuned on CVE-CWE pairs performs on par with larger LLMs for this task.
\citet{talibzade2025llms} find that for both CWE and severity prediction, TF-IDF features outperform larger LLMs prompted generically about the task, which they hypothesize is due to the highly schematic nature of CVE descriptions.
We find similar results on zero-shot severity prediction in Section~\ref{sec:findings}.

CVE-LLM~\citep{ghosh2025cvellm} fine-tunes on vulnerabilities in medical devices, using a domain ontology for supervision, and \citet{torkamani2025casey} use LLMs to predict CWE and severity as part of a vulnerability triage tool.

\paragraph{LLMs for vulnerability-related tasks.}
Recent work has considered LLMs on a range of vulnerability-related tasks.
VulnWatch~\citep{databricks2025vulnwatch} uses LLMs to prioritize vulnerabilities.
\citet{roytman2025benchmarking} finds that, for predicting likely exploitation (EPSS scores), even the best performing frontier LLMs lag behind state of the art statistical models, while costing orders of magnitude more to query at scale.

Unlike VulnWatch~\citep{databricks2025vulnwatch} and much previous work, we run our models on local hardware and evaluate on vulnerabilities introduced strictly after those in our training data.

\section{Limitations}\label{sec:limitations}

Our CWE evaluation using descriptions alone is likely to underestimate achievable performance, since CWE prediction is much easier when, unlike us but like NVD analysts, you can draw on context outside the description (e.g., by reading the original advisory).
This point bears out in the CWEs predicted incorrectly according to the NVD in Table~\ref{tab:examples}, where the CVSS vectors are nonetheless correct.
(More CWE disagreements between the model and the NVD appear in Section~\ref{sec:discussion}.)
For CVE-2024-0204 (an authentication bypass in GoAnywhere MFT), the NVD chooses CWE-425 (direct request, also called forced browsing), while our model chooses CWE-287 (improper authentication), arguably a more intuitive CWE.
For CVE-2024-0149 (a vulnerability in an NVIDIA driver allowing ``unauthorized access to files'' and a ``limited information disclosure''), the NVD chooses CWE-125 (out-of-bounds read), while our model chooses CWE-200 (exposure of sensitive information); it's true that the information is sensitive, but the description makes no mention of an out-of-bounds read.
Some fraction of the CWE errors reported in Table~\ref{tab:main} stem from such disagreements between the text and the NVD's CWE labels, and several of the other errors in Section~\ref{sec:discussion} are alternative interpretations a human might endorse.

Our benchmark evaluates each model on a single (held out) year's worth of test CVEs.
Moreover, each model is trained just once.
Accordingly, the error bars reported in Appendix~\ref{app:results} do not reflect variability from training, but only from randomly sampling test-year CVEs.

For space reasons, we only report the performance of our zero-shot LLM on a random sample of \nSub{} test CVEs, not on the entire year as for other models.
As such, the confidence intervals reported in Table~\ref{tab:subset} do not reflect the size of our full test set, but rather the \nSub{}-CVE subset.

\section{Conclusion}\label{sec:conclusion}

Our results suggest that CVE enrichment is tractable with modest computational resources, using only CVE description text.
Our TF-IDF logistic regression model trains in just a few minutes, requires no GPU, and trails the fine-tuned LLM by just a few points in each CVSS component and CWE.
Moreover, our simple formula scoring, which costs nothing, rescues the TF-IDF model's rare tiers; we therefore recommend deploying formula scoring on top of TF-IDF classification in all but applications that prize precision over recall on Critical vulnerabilities.
The 3.8B fine-tuned model trains in just \nLoraTrainHours{} hours, and achieves better results across the board: more consistent severity scores and CVSS vectors, and CWEs outside the frequent CWEs.
At inference time, it takes on the order of a half second to predict each CVE.
If either throughput or accuracy were insufficient, we might train for more than one epoch (validation accuracy was still climbing; see Table~\ref{tab:ckpt}), oversample or weight the loss of Low predictions (class-balanced weights did help our TF-IDF model), or try larger open models amenable to fine-tuning.
We leave these directions to future work.

We did not evaluate CVSS v4.0 vectors, which are starting to appear for new CVEs.
We expect formula scoring to work with these vectors too, though we have not tested it.
Our fine-tuned LLM could perhaps be retrained on the new format.

Our zero-shot LLM performs poorly by comparison.
We do not recommend using it for CVE enrichment.

Practitioners might wonder how best to use these models.
While we did not train or evaluate an ensemble, both our trained models make errors on mostly different CVEs; in principle, then, they might achieve up to \nEitherSev\% accuracy on tiers if ensembled somehow (e.g., by prediction agreement or model confidence).
An ensemble along these lines might or might not come close to this upper bound; we leave such investigations to future work.
Until then, we would be loath to deploy either trained model without bearing in mind that their Low tiers are unreliable, as is the privileges required component of CVSS vectors.
A CVE should be checked before a predicted 9.8 score moves it to the front of the triage queue.

\bibliographystyle{plainnat}
\bibliography{refs}

\appendix
\let\origtable\table
\let\endorigtable\endtable
\renewenvironment{table}[1][]{\origtable[H]}{\endorigtable}
\section{Prompts}\label{app:prompts}
Both prompts are sent as a single user turn through the Phi-3.5 chat template, and descriptions longer than 2,000 characters (\nTestClipped\% of test CVEs) are cut at that length.
\paragraph{Fine-tuned model (user turn).}
{\footnotesize \begin{verbatim}
You are a cybersecurity expert. Given the CVE description below, predict the
NVD enrichment fields.
Return ONLY a JSON object with no extra text.

CVE Description:
<CVE description>
\end{verbatim}}

\paragraph{Zero-shot model (user turn).}
{\footnotesize \begin{verbatim}
You are a cybersecurity expert specializing in vulnerability analysis. Given a
CVE description, predict the NVD CVSS v3.1 enrichment fields.

CVE Description:
<CVE description>

Respond ONLY with one JSON object with exactly these keys and allowed values:
{"attack_vector": "NETWORK"|"ADJACENT_NETWORK"|"LOCAL"|"PHYSICAL",
"attack_complexity": "LOW"|"HIGH", "privileges_required": "NONE"|"LOW"|"HIGH",
"user_interaction": "NONE"|"REQUIRED", "scope": "UNCHANGED"|"CHANGED",
"confidentiality": "NONE"|"LOW"|"HIGH", "integrity": "NONE"|"LOW"|"HIGH",
"availability": "NONE"|"LOW"|"HIGH", "cwe": "CWE-<number>", "cvss_score":
<number 0.0-10.0>, "severity": "CRITICAL"|"HIGH"|"MEDIUM"|"LOW"}
\end{verbatim}}

\section{Reproducibility}\label{app:repro}
All code, configuration files, and per-CVE predictions are in the project repository, the NVD snapshot is pinned to FKIE release \texttt{v2026.10.05-000016}, and each step below is a single command that regenerates its part of the paper.
\begin{itemize}\itemsep0pt
  \item Data preparation with \texttt{download\_data.py} and \texttt{prepare\_splits.py}.
  \item TF-IDF baselines with \texttt{baselines\_classical.py}.
  \item LoRA training with \texttt{mlx\_lm.lora} and \texttt{scripts/lora\_phi35.yaml}.
  \item Checkpoint selection with \texttt{select\_checkpoint.py}.
  \item Evaluation with \texttt{evaluate\_lora.py}.
  \item Tables, figures, and numbers in the text with \texttt{make\_tables.py} and \texttt{plot\_results.py}.
\end{itemize}

\FloatBarrier
\section{Data and Training Details}\label{app:data}

\FloatBarrier
\subsection{Data}
We use the publicly available National Vulnerability Database (NVD)\footnote{\url{https://nvd.nist.gov}} for vulnerability data.
Up until recently the NVD enriched nearly all CVEs with CWEs, CVSS base scores, severity labels and more, but since April 2026~\citep{nist2026nvd} it only enriches vulnerabilities it considers high-priority.

We include all CVEs with a CWE, primary\footnote{When a CVE has several CVSS entries, we use the primary score assigned by NIST.} base CVSS v3 score and severity, published from 2022 to 2024 (both included), for a total of \nAll.
Of these, \nVThreeZero{} have CVSS v3.0 scores, and the rest v3.1.
The formula for computing the base score from the eight components gives the same score in both versions for every one of these CVEs.

For training/validation/test we use a temporal split, to prevent data leakage and better reflect real use: We train and validate on older CVEs, and test on newer ones.
The \nTrainVal{} CVEs with identifiers from 2022 and 2023 are split into training and validation data randomly with an 80/20 ratio, stratified by severity, resulting in \nTrain{} training and \nVal{} validation CVEs, Table~\ref{tab:data}.
The test data are the \nTest{} remaining (i.e., usable) CVEs, all with identifiers from 2024.
Note that Low severity is less than 2\% of all CVEs, in all splits.
The CWEs exhibit a long tail, Table~\ref{tab:data}.
There are \nTrainCwes{} different CWEs in the training set and \nTestCwes{} in the test set, but the top 100 CWEs in the training set occur in \nTopCov\% of the test CVEs.
We do not use the recent VLAI data and model of \citet{bonhomme2025vlai} because the random train/validation/test split of the underlying dataset, \texttt{CIRCL/vulnerability-scores}, includes CVEs from 2024 (also present in our test data), 2025 and 2026, so VLAI's performance could in part reflect memorisation rather than generalisation ability.
A temporal split is necessary to fairly compare all systems' abilities to generalise, particularly since products mentioned in descriptions repeat often.

\FloatBarrier
\subsection{TF-IDF Baselines}
As baselines we implement TF-IDF classification/regression methods similar to previous work~\citep{elbaz2020fighting}.
We create TF-IDF representations with unigrams and bigrams, using sublinear tf and min document frequency 2, including stop words since words such as ``not'' or ``without'' can affect vulnerability meaning.
We limit the number of features to 50,000.
We train a different model for each output, Table~\ref{tab:summary}, using scikit-learn~\citep{pedregosa2011sklearn}:
\begin{description}\sloppy
  \item[LogReg] We use logistic regression, and select the regularisation strength, $C \in \{0.3, 1, 3, 10\}$, based on validation set accuracy, separately for severity and CWE classification, and for base component classification when predicting all eight components at once.
  \item[CB-LogReg] Class-balanced logistic regression: Identical setup, except we weight each example inversely proportional to its true class frequency.
  \item[Rand.\ Forest] Untuned random forest classifier, with 300 trees for severity and CWE, or 100 for each base component.
  \item[Ridge] Untuned ridge regression to predict scores, except that we choose the $\alpha \in \{0.3, 1, 3, 10\}$ with lowest MAE in validation.
\end{description}
For CWE we only train the models on the most frequent 100 CWEs in the training data, plus an \textsc{Other} class for the remaining CWEs.
A prediction of \textsc{Other} is always considered an error since we compare with the CVEs' true CWEs.

\FloatBarrier
\subsection{Language Model}
We use the 3.8B-parameter phi-3.5-mini-instruct (Phi) causal LM (decoder-only LLM)~\citep{abdin2024phi3}.
Unlike VulnWatch~\citep{databricks2025vulnwatch}, a system that utilises hosted (commercial) LLMs, Phi is available for download, and specialised enough, to be practically usable on local hardware.
Phi's training data is reported to end in October 2023,\footnote{\url{https://huggingface.co/microsoft/Phi-3.5-mini-instruct}} i.e., before any CVEs in the test set.

We fine-tune with LoRA~\citep{hu2022lora} using MLX~\citep{hannun2023mlx}.
The LoRAs (rank 16, scale 20, dropout 0.05) are applied to all linear projections in the final 16 transformer layers, totalling \nLoraParams M parameters, only \nLoraPct\% of Phi's parameters.
We train for 5,000 steps (one epoch) with batch size 8.
The learning rate is $10^{-4}$ for the first 100 (warm-up) steps, then decays cosinally to $10^{-5}$.
We exclude \nDropped{} examples (\nDroppedPct\%) with more than 768 tokens to limit memory use.
In contrast we include all test set CVEs, irrespective of length.

The fine-tuning target is a single JSON object: First the eight base score components, then CWE, score, and severity.
Placing score after components ensures that Phi's output is internally consistent, since it has already (essentially)\footnote{Generation is not constrained, so the model can still write a score that disagrees with its vector. It does so on less than 1\% of test CVEs (Table~\ref{tab:consistency}).} committed to a vector by the time it outputs the score, similar to our formula scoring method (Section~\ref{sec:evaluation}).
We evaluate a saved model every 1,000 training steps on a sample of 1,000 validation CVEs, scoring predicted scores via formula, and keep the one with best severity accuracy using greedy decoding, Table~\ref{tab:ckpt}.

For zero-shot Phi we use a prompt with a brief task description, all output field names and the legal values for components and severity (Appendix~\ref{app:prompts}).
Zero-shot Phi is very verbose and slow, so to limit run time we only include it on a random sample of \nSub{} CVEs from the test set.
Where relevant, we evaluate other systems on the same sample.

\FloatBarrier
\subsection{Dataset Statistics}\label{app:stats}
\begin{table}[t]
\centering
\caption{Data splits, with severity columns giving the share of CVEs (\%) in each tier.}
\label{tab:data}
\small
\begin{tabular}{lrrrrrr}
\toprule
Split & CVEs & Critical & High & Medium & Low & CWEs \\
\midrule
Train (2022--23) & 40{,}436 & 15.4 & 39.3 & 43.7 & 1.7 & 464 \\
Validation (2022--23) & 10{,}109 & 15.4 & 39.3 & 43.7 & 1.7 & 293 \\
Test (2024) & 36{,}325 & 13.2 & 36.2 & 48.6 & 1.9 & 512 \\
\bottomrule
\end{tabular}
\end{table}

\section{Full Results}\label{app:results}

\FloatBarrier
\subsection{All Systems}
\begin{table}[t]
\centering
\caption{All systems on the full 2024 test set (36{,}325 CVEs). Values are percentages except score MAE. $\pm$ gives the half-width of the 95\% bootstrap interval. Bold marks the best value in each column. The TF-IDF direct systems share one ridge regression, so their score MAE is identical.}
\label{tab:main}
\small
\setlength{\tabcolsep}{4pt}
\begin{tabular}{lcccccc}
\toprule
 & \multicolumn{2}{c}{Severity} & CWE & Score & \multicolumn{2}{c}{Vector} \\
\cmidrule(lr){2-3}\cmidrule(lr){6-7}
System & Acc. & Macro-F1 & Acc. & MAE $\downarrow$ & Partial & Exact \\
\midrule
TF-IDF + LR, direct & 67.1\,{\scriptsize$\pm$0.5} & 47.9\,{\scriptsize$\pm$0.6} & 67.2\,{\scriptsize$\pm$0.5} & 0.98\,{\scriptsize$\pm$0.01} & 86.9\,{\scriptsize$\pm$0.2} & 44.1\,{\scriptsize$\pm$0.5} \\
TF-IDF + LR, formula & 67.1\,{\scriptsize$\pm$0.5} & 51.6\,{\scriptsize$\pm$0.8} & 67.2\,{\scriptsize$\pm$0.5} & 0.94\,{\scriptsize$\pm$0.01} & 86.9\,{\scriptsize$\pm$0.2} & 44.1\,{\scriptsize$\pm$0.5} \\
TF-IDF + LR balanced, direct & 65.7\,{\scriptsize$\pm$0.5} & 51.9\,{\scriptsize$\pm$0.8} & 67.2\,{\scriptsize$\pm$0.5} & 0.98\,{\scriptsize$\pm$0.01} & 86.0\,{\scriptsize$\pm$0.2} & 40.5\,{\scriptsize$\pm$0.5} \\
TF-IDF + LR balanced, formula & 67.1\,{\scriptsize$\pm$0.5} & 53.9\,{\scriptsize$\pm$0.9} & 67.2\,{\scriptsize$\pm$0.5} & 0.98\,{\scriptsize$\pm$0.01} & 86.0\,{\scriptsize$\pm$0.2} & 40.5\,{\scriptsize$\pm$0.5} \\
TF-IDF + RF, direct & 66.9\,{\scriptsize$\pm$0.5} & 47.1\,{\scriptsize$\pm$0.6} & 65.2\,{\scriptsize$\pm$0.5} & 0.98\,{\scriptsize$\pm$0.01} & 86.1\,{\scriptsize$\pm$0.2} & 41.8\,{\scriptsize$\pm$0.5} \\
TF-IDF + RF, formula & 64.0\,{\scriptsize$\pm$0.5} & 47.8\,{\scriptsize$\pm$0.6} & 65.2\,{\scriptsize$\pm$0.5} & 1.03\,{\scriptsize$\pm$0.01} & 86.1\,{\scriptsize$\pm$0.2} & 41.8\,{\scriptsize$\pm$0.5} \\
Phi-3.5 + LoRA, direct & \textbf{69.9}\,{\scriptsize$\pm$0.5} & \textbf{55.7}\,{\scriptsize$\pm$0.9} & \textbf{70.9}\,{\scriptsize$\pm$0.5} & \textbf{0.85}\,{\scriptsize$\pm$0.01} & \textbf{87.9}\,{\scriptsize$\pm$0.2} & \textbf{48.8}\,{\scriptsize$\pm$0.5} \\
Phi-3.5 + LoRA, formula & \textbf{69.9}\,{\scriptsize$\pm$0.5} & \textbf{55.7}\,{\scriptsize$\pm$0.9} & \textbf{70.9}\,{\scriptsize$\pm$0.5} & \textbf{0.85}\,{\scriptsize$\pm$0.01} & \textbf{87.9}\,{\scriptsize$\pm$0.2} & \textbf{48.8}\,{\scriptsize$\pm$0.5} \\
\bottomrule
\end{tabular}
\end{table}

\FloatBarrier
\subsection{Paired Differences}
\begin{table}[h]
\centering
\caption{Fine-tuned Phi-3.5 minus TF-IDF logistic regression with formula scoring, on the same \nTest{} test CVEs, with 95\% paired bootstrap intervals.}
\label{tab:paired}
\small
\begin{tabular}{lcc}
\toprule
Metric & Difference & 95\% interval \\
\midrule
Severity accuracy (points) & \nDiffLRfSev & [\nDiffLRfSevLo, \nDiffLRfSevHi] \\
Severity macro-F1 (points) & \nDiffLRfFone & [\nDiffLRfFoneLo, \nDiffLRfFoneHi] \\
CWE accuracy (points) & \nDiffLRfCwe & [\nDiffLRfCweLo, \nDiffLRfCweHi] \\
Score MAE & \nDiffLRfMae & [\nDiffLRfMaeLo, \nDiffLRfMaeHi] \\
Exact CVSS vector (points) & \nDiffLRfVec & [\nDiffLRfVecLo, \nDiffLRfVecHi] \\
\bottomrule
\end{tabular}
\end{table}

\FloatBarrier
\subsection{Component Values}
\begin{table}[t]
\centering
\caption{Share of 2024 test CVEs (\%) with each value of each CVSS v3.1 component, as assigned by NIST and as predicted by the fine-tuned model.}
\label{tab:valuefreq}
\small
\begin{tabular}{llrr}
\toprule
Component & Value & NVD & Fine-tuned \\
\midrule
Attack vector & Network & 76.2 & 79.2 \\
 & Adjacent & 2.9 & 0.9 \\
 & Local & 20.1 & 19.5 \\
 & Physical & 0.7 & 0.4 \\
\addlinespace[2pt]
Attack complexity & Low & 94.3 & 98.2 \\
 & High & 5.7 & 1.8 \\
\addlinespace[2pt]
Privileges required & Low & 40.3 & 35.2 \\
 & High & 9.4 & 4.7 \\
 & None & 50.4 & 60.1 \\
\addlinespace[2pt]
User interaction & None & 66.1 & 64.1 \\
 & Required & 33.9 & 35.9 \\
\addlinespace[2pt]
Scope & Unchanged & 73.5 & 76.6 \\
 & Changed & 26.5 & 23.4 \\
\addlinespace[2pt]
Confidentiality & Low & 30.6 & 25.2 \\
 & High & 48.8 & 56.0 \\
 & None & 20.6 & 18.8 \\
\addlinespace[2pt]
Integrity & Low & 32.1 & 27.8 \\
 & High & 41.2 & 45.9 \\
 & None & 26.7 & 26.3 \\
\addlinespace[2pt]
Availability & Low & 9.0 & 0.2 \\
 & High & 48.8 & 55.3 \\
 & None & 42.2 & 44.5 \\
\bottomrule
\end{tabular}
\end{table}

\FloatBarrier
\subsection{Exact CVSS Vectors}\label{app:exact}
\begin{table}[t]
\centering
\caption{Which component is wrong when exactly one of the eight is wrong (\% of such CVEs), for the fine-tuned model and TF-IDF logistic regression on the 2024 test set.}
\label{tab:sole}
\small
\begin{tabular}{lcc}
\toprule
Component & Fine-tuned & TF-IDF \\
\midrule
Attack vector & 7 & 7 \\
Attack complexity & 6 & 4 \\
Privileges required & 38 & 38 \\
User interaction & 12 & 10 \\
Scope & 5 & 5 \\
Confidentiality & 10 & 12 \\
Integrity & 6 & 8 \\
Availability & 17 & 16 \\
\bottomrule
\end{tabular}
\end{table}

\begin{table}[t]
\centering
\caption{CVEs whose severity is predicted correctly, split by how many of the eight vector components are wrong (\% of all 2024 test CVEs). Severity can be right while the vector is not, and most such CVEs have exactly one wrong component.}
\label{tab:gap}
\small
\begin{tabular}{lccccc}
\toprule
 & Severity correct & 0 wrong & 1 wrong & 2 wrong & 3+ wrong \\
\midrule
Fine-tuned Phi-3.5 & 69.9 & 48.8 & 13.1 & 4.9 & 3.2 \\
TF-IDF + LR, formula & 67.1 & 44.1 & 14.6 & 5.3 & 3.1 \\
\bottomrule
\end{tabular}
\end{table}

\FloatBarrier
\section{Additional Analysis}\label{app:analysis}

\begin{table}[t]
\centering
\caption{Test CVEs discussed in Section~\ref{sec:discussion}, with the NVD labels and the output of the fine-tuned model. The last column lists the vector components the model got wrong.}
\label{tab:examples}
\footnotesize
\setlength{\tabcolsep}{4pt}
\begin{tabular}{@{}lp{5.3cm}p{1.9cm}p{1.9cm}l@{}}
\toprule
CVE & Description & NVD & Predicted & Wrong \\
\midrule
\texttt{CVE-2024-0003} & A condition exists in FlashArray Purity whereby a malicious user could use a remote administrative service to create an account on the array allowing privileged access. & 7.2 High, CWE-269 & 9.8 Critical, CWE-287 & PR \\
\texttt{CVE-2024-0008} & Web sessions in the management interface in Palo Alto Networks PAN-OS software do not expire in certain situations, making it susceptible to unauthorized access. & 8.8 High, CWE-613 & 9.8 Critical, CWE-613 & UI \\
\texttt{CVE-2024-0257} & RoboDK v5.5.4 is vulnerable to heap-based buffer overflow while processing a specific project file. The resulting memory corruption may crash the application. & 3.3 Low, CWE-122 & 7.8 High, CWE-787 & C, I, A \\
\texttt{CVE-2024-10214} & Mattermost versions 9.11.X \textless{}= 9.11.1, 9.5.x \textless{}= 9.5.9 icorrectly issues two sessions when using desktop SSO - one in the browser and one in desktop with incorrect settings. & 3.5 Low, CWE-303 & 9.8 Critical, CWE-287 & PR, UI, C, I, A \\
\texttt{CVE-2024-0204} & Authentication bypass in Fortra's GoAnywhere MFT prior to 7.4.1 allows an unauthorized user to create an admin user via the administration portal. & 9.8 Critical, CWE-425 & 9.8 Critical, CWE-287 & none \\
\texttt{CVE-2024-0149} & NVIDIA GPU Display Driver for Linux contains a vulnerability which could allow an attacker unauthorized access to files. A successful exploit of this vulnerability might lead to limited information disclosure. & 3.3 Low, CWE-125 & 3.3 Low, CWE-200 & none \\
\bottomrule
\end{tabular}
\end{table}

\begin{table}[t]
\centering
\caption{Mean absolute score error by true severity tier on the 2024 test set.}
\label{tab:tiermae}
\small
\begin{tabular}{lcccc}
\toprule
System & Critical & High & Medium & Low \\
\midrule
TF-IDF ridge regression (direct) & 1.23 & 0.80 & 0.96 & 2.91 \\
TF-IDF + LR, formula & 0.59 & 0.81 & 1.04 & 3.26 \\
Phi-3.5 + LoRA, formula & 0.53 & 0.76 & 0.92 & 2.90 \\
\bottomrule
\end{tabular}
\end{table}

\begin{table}[t]
\centering
\caption{Severity accuracy and exact-vector rate (\%) by quartile of description length on the 2024 test set. Quartile boundaries are 194, 285, and 444 characters.}
\label{tab:length}
\small
\setlength{\tabcolsep}{4pt}
\begin{tabular}{lcccccccc}
\toprule
 & \multicolumn{2}{c}{Shortest quartile} & \multicolumn{2}{c}{Second} & \multicolumn{2}{c}{Third} & \multicolumn{2}{c}{Longest quartile} \\
\cmidrule(lr){2-3}\cmidrule(lr){4-5}\cmidrule(lr){6-7}\cmidrule(lr){8-9}
System & Sev. & Vector & Sev. & Vector & Sev. & Vector & Sev. & Vector \\
\midrule
TF-IDF + LR, formula & 59.7 & 38.8 & 67.2 & 38.4 & 71.0 & 50.7 & 70.6 & 48.7 \\
Phi-3.5 + LoRA, formula & 61.7 & 42.0 & 69.7 & 42.0 & 74.4 & 55.7 & 73.9 & 55.5 \\
\bottomrule
\end{tabular}
\end{table}

\begin{table}[t]
\centering
\caption{Severity tiers on the 2{,}000-CVE zero-shot sample: as assigned by NIST, as written by zero-shot Phi-3.5, and as computed from the vector it wrote. Counts with shares (\%) in parentheses.}
\label{tab:zssev}
\small
\begin{tabular}{lccc}
\toprule
Tier & NVD & Zero-shot, written & Zero-shot, from vector \\
\midrule
Critical & 266 (13.3) & 1{,}247 (62.4) & 68 (3.4) \\
High & 742 (37.1) & 697 (34.8) & 558 (27.9) \\
Medium & 951 (47.5) & 9 (0.4) & 1{,}146 (57.3) \\
Low & 41 (2.1) & 2 (0.1) & 15 (0.8) \\
No valid tier & -- & 45 (2.2) & -- \\
\bottomrule
\end{tabular}
\end{table}

\begin{table}[t]
\centering
\caption{Validation metrics of the fine-tuned model under formula scoring at each saved checkpoint, on the same 1,000 validation CVEs.}
\label{tab:ckpt}
\small
\begin{tabular}{rccccc}
\toprule
Step & Severity acc. & Macro-F1 & CWE acc. & Score MAE & Exact vector \\
\midrule
1{,}000 & 68.2 & 50.8 & 65.4 & 0.83 & 52.9 \\
2{,}000 & 72.6 & 56.9 & 67.2 & 0.73 & 55.4 \\
3{,}000 & 74.4 & 58.5 & 71.1 & 0.66 & 62.2 \\
4{,}000 & 76.0 & 62.1 & 73.4 & 0.65 & 63.0 \\
5{,}000 & 76.3 & 58.6 & 74.2 & 0.63 & 63.9 \\
\bottomrule
\end{tabular}
\end{table}

\begin{table}[htbp]
\centering
\caption{Wall-clock time on a single machine, with inference run in small batches of at most 32 CVEs and without optimization for speed.}
\label{tab:compute}
\small
\begin{tabular}{lr}
\toprule
Step & Time \\
\midrule
All TF-IDF baselines, tuning and test (CPU) & 9 min \\
LoRA fine-tuning, 5{,}000 steps & 2.5 h \\
LoRA inference, 36{,}325 test CVEs & 4.9 h \\
\bottomrule
\end{tabular}
\end{table}

\FloatBarrier
\begin{table}[p]
\centering
\caption{Test CVEs on which all four trained systems predicted the wrong value for the listed component (attack vector, attack complexity, privileges required, and user interaction). Bold marks words in the NVD description that may signal the true value. FT: fine-tuned Phi-3.5, LR: logistic regression, CB: class-balanced logistic regression, RF: random forest, all with formula scoring. Values: N network or none, A adjacent, L local or low, P physical, H high, R required, U unchanged, C changed.}
\label{tab:qualalla}
\scriptsize
\setlength{\tabcolsep}{3pt}
\begin{tabular}{@{}llp{6.2cm}cl@{}}
\toprule
Component & CVE & Description & True & Predicted \\
\midrule
Attack vector & \texttt{CVE-2024-0001} & A condition exists in FlashArray Purity whereby a local account intended for initial array configuration remains active potentially allowing a malicious actor to gain elevated privileges. & N & FT:L LR:L CB:L RF:L \\
 & \texttt{CVE-2024-0311} & A malicious insider can bypass the existing policy of Skyhigh Client Proxy without a valid release code. & L & FT:N LR:N CB:N RF:N \\
 & \texttt{CVE-2024-0315} & Remote file inclusion vulnerability in FireEye Central Management affecting version 9.1.1.956704. This vulnerability allows an attacker to upload a malicious PDF file to the system during the report creation process. & L & FT:N LR:N CB:N RF:N \\
 & \texttt{CVE-2024-0321} & Stack-based Buffer Overflow in GitHub repository gpac/gpac prior to 2.3-DEV. & N & FT:L LR:L CB:L RF:L \\
\addlinespace[3pt]
Attack complexity & \texttt{CVE-2024-0056} & Microsoft.Data.SqlClient and System.Data.SqlClient SQL Data Provider Security Feature Bypass Vulnerability & H & FT:L LR:L CB:L RF:L \\
 & \texttt{CVE-2024-0144} & NVIDIA nvJPEG2000 library contains a vulnerability where an attacker can cause a buffer overflow issue by means of a specially crafted JPEG2000 file. A successful exploit of this vulnerability might lead to data tampering. & H & FT:L LR:L CB:L RF:L \\
 & \texttt{CVE-2024-0410} & An authorization bypass vulnerability was discovered in GitLab affecting versions 15.1 prior to 16.7.6, 16.8 prior to 16.8.3, and 16.9 prior to 16.9.1. A developer could bypass CODEOWNERS approvals by creating a merge conflict. & H & FT:L LR:L CB:L RF:L \\
 & \texttt{CVE-2024-0980} & The Auto-update service for Okta Verify for Windows is vulnerable to two flaws which in combination could be used to execute arbitrary code. & H & FT:L LR:L CB:L RF:L \\
\addlinespace[3pt]
Privileges required & \texttt{CVE-2024-0179} & SMM Callout vulnerability within the AmdCpmDisplayFeatureSMM driver could allow locally authenticated attackers to overwrite SMRAM, potentially resulting in arbitrary code execution. & H & FT:L LR:L CB:L RF:L \\
 & \texttt{CVE-2024-0231} & A resource misdirection vulnerability in GitLab CE/EE versions 12.0 prior to 17.0.5, 17.1 prior to 17.1.3, and 17.2 prior to 17.2.1 allows an attacker to craft a repository import in such a way as to misdirect commits. & H & FT:L LR:N CB:N RF:N \\
 & \texttt{CVE-2024-0311} & A malicious insider can bypass the existing policy of Skyhigh Client Proxy without a valid release code. & L & FT:N LR:N CB:N RF:N \\
 & \texttt{CVE-2024-0315} & Remote file inclusion vulnerability in FireEye Central Management affecting version 9.1.1.956704. This vulnerability allows an attacker to upload a malicious PDF file to the system during the report creation process. & L & FT:N LR:N CB:N RF:N \\
\addlinespace[3pt]
User interaction & \texttt{CVE-2024-0008} & Web sessions in the management interface in Palo Alto Networks PAN-OS software do not expire in certain situations, making it susceptible to unauthorized access. & R & FT:N LR:N CB:N RF:N \\
 & \texttt{CVE-2024-0321} & Stack-based Buffer Overflow in GitHub repository gpac/gpac prior to 2.3-DEV. & N & FT:R LR:R CB:R RF:R \\
 & \texttt{CVE-2024-0322} & Out-of-bounds Read in GitHub repository gpac/gpac prior to 2.3-DEV. & N & FT:R LR:R CB:R RF:R \\
 & \texttt{CVE-2024-0410} & An authorization bypass vulnerability was discovered in GitLab affecting versions 15.1 prior to 16.7.6, 16.8 prior to 16.8.3, and 16.9 prior to 16.9.1. A developer could bypass CODEOWNERS approvals by creating a merge conflict. & R & FT:N LR:N CB:N RF:N \\
\bottomrule
\end{tabular}
\end{table}

\begin{table}[p]
\centering
\caption{Test CVEs on which all four trained systems predicted the wrong value for the listed component (scope and the three impacts). Bold marks words in the NVD description that may signal the true value. FT: fine-tuned Phi-3.5, LR: logistic regression, CB: class-balanced logistic regression, RF: random forest, all with formula scoring. Values: N network or none, A adjacent, L local or low, P physical, H high, R required, U unchanged, C changed.}
\label{tab:qualallb}
\scriptsize
\setlength{\tabcolsep}{3pt}
\begin{tabular}{@{}llp{6.2cm}cl@{}}
\toprule
Component & CVE & Description & True & Predicted \\
\midrule
Scope & \texttt{CVE-2024-0056} & Microsoft.Data.SqlClient and System.Data.SqlClient SQL Data Provider Security Feature Bypass Vulnerability & C & FT:U LR:U CB:U RF:U \\
 & \texttt{CVE-2024-0171} & Dell PowerEdge Server BIOS contains an TOCTOU race condition vulnerability. A local low privileged attacker could potentially exploit this vulnerability to gain access to otherwise unauthorized resources. & C & FT:U LR:U CB:U RF:U \\
 & \texttt{CVE-2024-0179} & SMM Callout vulnerability within the AmdCpmDisplayFeatureSMM driver could allow locally authenticated attackers to overwrite SMRAM, potentially resulting in arbitrary code execution. & C & FT:U LR:U CB:U RF:U \\
 & \texttt{CVE-2024-0410} & An authorization bypass vulnerability was discovered in GitLab affecting versions 15.1 prior to 16.7.6, 16.8 prior to 16.8.3, and 16.9 prior to 16.9.1. A developer could bypass CODEOWNERS approvals by creating a merge conflict. & C & FT:U LR:U CB:U RF:U \\
\addlinespace[3pt]
Confidentiality & \texttt{CVE-2024-0171} & Dell PowerEdge Server BIOS contains an TOCTOU race condition vulnerability. A local low privileged attacker could potentially exploit this vulnerability to gain access to otherwise unauthorized resources. & L & FT:H LR:H CB:H RF:H \\
 & \texttt{CVE-2024-0257} & RoboDK v5.5.4 is vulnerable to heap-based buffer overflow while processing a specific project file. The resulting memory corruption may \textbf{crash} the application. & N & FT:H LR:H CB:H RF:H \\
 & \texttt{CVE-2024-0403} & Recipes version 1.5.10 allows arbitrary HTTP requests to be made through the server. This is possible because the application is vulnerable to SSRF. & L & FT:H LR:H CB:N RF:H \\
 & \texttt{CVE-2024-0677} & The Pz-LinkCard WordPress plugin through 2.5.1 does not prevent users from pinging arbitrary hosts via some of its shortcodes, which could allow high privilege users such as contributors to perform SSRF attacks. & L & FT:H LR:H CB:H RF:H \\
\addlinespace[3pt]
Integrity & \texttt{CVE-2024-0171} & Dell PowerEdge Server BIOS contains an TOCTOU race condition vulnerability. A local low privileged attacker could potentially exploit this vulnerability to gain access to otherwise unauthorized resources. & L & FT:H LR:H CB:H RF:H \\
 & \texttt{CVE-2024-0257} & RoboDK v5.5.4 is vulnerable to heap-based buffer overflow while processing a specific project file. The resulting memory corruption may \textbf{crash} the application. & N & FT:H LR:H CB:H RF:H \\
 & \texttt{CVE-2024-0741} & An out of bounds write in ANGLE could have allowed an attacker to corrupt memory leading to a potentially exploitable \textbf{crash}. This vulnerability affects Firefox \textless{} 122, Firefox ESR \textless{} 115.7, and Thunderbird \textless{} 115.7. & N & FT:H LR:H CB:H RF:H \\
 & \texttt{CVE-2024-0749} & A phishing site could have repurposed an `about:` dialog to show phishing content with an incorrect origin in the address bar. This vulnerability affects Firefox \textless{} 122 and Thunderbird \textless{} 115.7. & L & FT:H LR:H CB:H RF:H \\
\addlinespace[3pt]
Availability & \texttt{CVE-2024-0009} & An improper verification vulnerability in the GlobalProtect gateway feature of Palo Alto Networks PAN-OS software enables a malicious user with stolen credentials to establish a VPN connection from an unauthorized IP address. & L & FT:H LR:N CB:N RF:N \\
 & \texttt{CVE-2024-0068} & Improper Link Resolution Before File Access ('Link Following') vulnerability in HYPR Workforce Access on MacOS allows File Manipulation.This issue affects Workforce Access: before 8.7.1. & N & FT:H LR:H CB:H RF:H \\
 & \texttt{CVE-2024-0160} & Dell Client Platform contains an incorrect authorization vulnerability. An attacker with physical access to the system could potentially exploit this vulnerability by bypassing BIOS authorization to modify settings in the BIOS. & H & FT:N LR:N CB:N RF:N \\
 & \texttt{CVE-2024-0171} & Dell PowerEdge Server BIOS contains an TOCTOU race condition vulnerability. A local low privileged attacker could potentially exploit this vulnerability to gain access to otherwise unauthorized resources. & L & FT:H LR:H CB:H RF:H \\
\bottomrule
\end{tabular}
\end{table}

\begin{table}[p]
\centering
\caption{Test CVEs on which at least one, but not all, of the four trained systems predicted the wrong value for the listed component (attack vector, attack complexity, privileges required, and user interaction). Bold marks words in the NVD description that may signal the true value. FT: fine-tuned Phi-3.5, LR: logistic regression, CB: class-balanced logistic regression, RF: random forest, all with formula scoring. Values: N network or none, A adjacent, L local or low, P physical, H high, R required, U unchanged, C changed.}
\label{tab:qualsomea}
\scriptsize
\setlength{\tabcolsep}{3pt}
\begin{tabular}{@{}llp{6.2cm}cl@{}}
\toprule
Component & CVE & Description & True & Predicted \\
\midrule
Attack vector & \texttt{CVE-2024-0092} & NVIDIA GPU Driver for Windows and Linux contains a vulnerability where an improper check or improper handling of exception conditions might lead to denial of service. & L & FT:L LR:L CB:P RF:L \\
 & \texttt{CVE-2024-0100} & NVIDIA Triton Inference Server for Linux contains a vulnerability in the tracing API, where a user can corrupt system files. A successful exploit of this vulnerability might lead to denial of service and data tampering. & N & FT:N LR:N CB:L RF:N \\
 & \texttt{CVE-2024-0144} & NVIDIA nvJPEG2000 library contains a vulnerability where an attacker can cause a buffer overflow issue by means of a specially crafted JPEG2000 file. A successful exploit of this vulnerability might lead to data tampering. & N & FT:L LR:L CB:L RF:N \\
 & \texttt{CVE-2024-0147} & NVIDIA GPU display driver for Windows and Linux contains a vulnerability where referencing memory after it has been freed can lead to denial of service or data tampering. & L & FT:L LR:L CB:P RF:L \\
\addlinespace[3pt]
Attack complexity & \texttt{CVE-2024-0057} & NET, .NET Framework, and Visual Studio Security Feature Bypass Vulnerability & L & FT:L LR:L CB:H RF:L \\
 & \texttt{CVE-2024-0179} & SMM Callout vulnerability within the AmdCpmDisplayFeatureSMM driver could allow locally authenticated attackers to overwrite SMRAM, potentially resulting in arbitrary code execution. & L & FT:L LR:L CB:H RF:L \\
 & \texttt{CVE-2024-10253} & A potential TOCTOU vulnerability was reported in PC Manager, Lenovo Browser, and Lenovo App Store that could allow a local attacker to cause a system crash. & H & FT:H LR:L CB:L RF:H \\
 & \texttt{CVE-2024-10397} & A malicious server can crash the OpenAFS cache manager and other client utilities, and possibly execute arbitrary code. & L & FT:L LR:L CB:H RF:L \\
\addlinespace[3pt]
Privileges required & \texttt{CVE-2024-0001} & A condition exists in FlashArray Purity whereby a local account intended for initial array configuration remains active potentially allowing a malicious actor to gain elevated privileges. & N & FT:L LR:N CB:N RF:L \\
 & \texttt{CVE-2024-0002} & A condition exists in FlashArray Purity whereby an attacker can employ a privileged account allowing remote access to the array. & N & FT:H LR:N CB:N RF:N \\
 & \texttt{CVE-2024-0003} & A condition exists in FlashArray Purity whereby a malicious user could use a remote \textbf{administrative} service to create an account on the array allowing privileged access. & H & FT:N LR:L CB:H RF:L \\
 & \texttt{CVE-2024-0004} & A condition exists in FlashArray Purity whereby an user with array \textbf{admin} role can execute arbitrary commands remotely to escalate privilege on the array. & H & FT:H LR:H CB:H RF:L \\
\addlinespace[3pt]
User interaction & \texttt{CVE-2024-0057} & NET, .NET Framework, and Visual Studio Security Feature Bypass Vulnerability & N & FT:N LR:R CB:R RF:N \\
 & \texttt{CVE-2024-0144} & NVIDIA nvJPEG2000 library contains a vulnerability where an attacker can cause a buffer overflow issue by means of a specially crafted JPEG2000 file. A successful exploit of this vulnerability might lead to data tampering. & R & FT:R LR:N CB:N RF:N \\
 & \texttt{CVE-2024-0207} & HTTP3 dissector crash in Wireshark 4.2.0 allows denial of service via packet injection or crafted capture file & N & FT:R LR:R CB:R RF:N \\
 & \texttt{CVE-2024-0208} & GVCP dissector crash in Wireshark 4.2.0, 4.0.0 to 4.0.11, and 3.6.0 to 3.6.19 allows denial of service via packet injection or crafted capture file & N & FT:R LR:R CB:R RF:N \\
\bottomrule
\end{tabular}
\end{table}

\begin{table}[p]
\centering
\caption{Test CVEs on which at least one, but not all, of the four trained systems predicted the wrong value for the listed component (scope and the three impacts). Bold marks words in the NVD description that may signal the true value. FT: fine-tuned Phi-3.5, LR: logistic regression, CB: class-balanced logistic regression, RF: random forest, all with formula scoring. Values: N network or none, A adjacent, L local or low, P physical, H high, R required, U unchanged, C changed.}
\label{tab:qualsomeb}
\scriptsize
\setlength{\tabcolsep}{3pt}
\begin{tabular}{@{}llp{6.2cm}cl@{}}
\toprule
Component & CVE & Description & True & Predicted \\
\midrule
Scope & \texttt{CVE-2024-0078} & NVIDIA GPU Display Driver for Windows and Linux contains a vulnerability in the kernel mode layer, where a user in a \textbf{guest} can cause a NULL-pointer dereference in the host, which may lead to denial of service. & C & FT:U LR:U CB:C RF:U \\
 & \texttt{CVE-2024-0315} & Remote file inclusion vulnerability in FireEye Central Management affecting version 9.1.1.956704. This vulnerability allows an attacker to upload a malicious PDF file to the system during the report creation process. & U & FT:U LR:U CB:C RF:U \\
 & \texttt{CVE-2024-0427} & The ARForms - Premium WordPress Form Builder Plugin WordPress plugin before 6.4.1 does not properly escape user-controlled input when it is reflected in some of its AJAX actions. & U & FT:C LR:U CB:U RF:U \\
 & \texttt{CVE-2024-0440} & Attacker, with permission to submit a link or submits a link via POST to be collected that is using the file:// protocol can then introspect host files and other relatively stored files. & U & FT:U LR:U CB:C RF:U \\
\addlinespace[3pt]
Confidentiality & \texttt{CVE-2024-0008} & Web sessions in the management interface in Palo Alto Networks PAN-OS software do not expire in certain situations, making it susceptible to unauthorized access. & H & FT:H LR:N CB:N RF:H \\
 & \texttt{CVE-2024-0009} & An improper verification vulnerability in the GlobalProtect gateway feature of Palo Alto Networks PAN-OS software enables a malicious user with stolen credentials to establish a VPN connection from an unauthorized IP address. & L & FT:H LR:N CB:L RF:H \\
 & \texttt{CVE-2024-0057} & NET, .NET Framework, and Visual Studio Security Feature Bypass Vulnerability & H & FT:H LR:H CB:L RF:H \\
 & \texttt{CVE-2024-0074} & NVIDIA GPU Display Driver for Linux contains a vulnerability where an attacker may access a memory location after the end of the buffer. A successful exploit of this vulnerability may lead to \textbf{denial of service} and data tampering. & N & FT:H LR:N CB:N RF:N \\
\addlinespace[3pt]
Integrity & \texttt{CVE-2024-0008} & Web sessions in the management interface in Palo Alto Networks PAN-OS software do not expire in certain situations, making it susceptible to unauthorized access. & H & FT:H LR:N CB:N RF:H \\
 & \texttt{CVE-2024-0009} & An improper verification vulnerability in the GlobalProtect gateway feature of Palo Alto Networks PAN-OS software enables a malicious user with stolen credentials to establish a VPN connection from an unauthorized IP address. & L & FT:H LR:L CB:L RF:H \\
 & \texttt{CVE-2024-0056} & Microsoft.Data.SqlClient and System.Data.SqlClient SQL Data Provider Security Feature Bypass Vulnerability & H & FT:H LR:N CB:N RF:H \\
 & \texttt{CVE-2024-0057} & NET, .NET Framework, and Visual Studio Security Feature Bypass Vulnerability & H & FT:H LR:L CB:L RF:H \\
\addlinespace[3pt]
Availability & \texttt{CVE-2024-0008} & Web sessions in the management interface in Palo Alto Networks PAN-OS software do not expire in certain situations, making it susceptible to unauthorized access. & H & FT:H LR:N CB:N RF:N \\
 & \texttt{CVE-2024-0056} & Microsoft.Data.SqlClient and System.Data.SqlClient SQL Data Provider Security Feature Bypass Vulnerability & N & FT:H LR:N CB:N RF:H \\
 & \texttt{CVE-2024-0057} & NET, .NET Framework, and Visual Studio Security Feature Bypass Vulnerability & H & FT:H LR:N CB:L RF:H \\
 & \texttt{CVE-2024-0144} & NVIDIA nvJPEG2000 library contains a vulnerability where an attacker can cause a buffer overflow issue by means of a specially crafted JPEG2000 file. A successful exploit of this vulnerability might lead to data tampering. & H & FT:N LR:H CB:H RF:H \\
\bottomrule
\end{tabular}
\end{table}

\end{document}